%% file: main.tex
\documentclass[10pt,a4paper,twocolumn]{article}

\usepackage[utf8]{inputenc}
\usepackage[T1]{fontenc}
\usepackage[english]{babel}
\usepackage{lmodern}
\usepackage{graphicx}
\usepackage{amsmath,amssymb}
\usepackage{booktabs}
\usepackage{xcolor}
\usepackage[hidelinks]{hyperref}
\usepackage[margin=2cm]{geometry}
\usepackage{microtype}

\providecommand{\tabwidth}{\columnwidth}
\providecommand{\tabwidthnarrow}{\columnwidth}
\providecommand{\hwjetsonwidth}{1.0\linewidth}
\providecommand{\archwidth}{0.75\textwidth}
\providecommand{\conceptwidth}{\linewidth}

\newif\ifsidebyside \sidebysidefalse
\usepackage{multicol}
\ifsidebyside\fi 
\AtBeginDocument{%
  \ifsidebyside
    \let\oldthebibliography\thebibliography
    \let\oldendthebibliography\endthebibliography
    \renewenvironment{thebibliography}[1]{%
      \begin{multicols}{2}\oldthebibliography{#1}%
    }{%
      \oldendthebibliography\end{multicols}%
    }%
  \fi
}

\usepackage[numbers,sort&compress]{natbib}
\graphicspath{{figures/}}

\title{Embedded Bi-Temporal Building Damage Assessment for
On-Board Data Reduction%
\thanks{Accepted at OBPDC 2026, Barcelona, October 2026. This is the authors'
preprint version. \copyright~2026 the authors, licensed under CC~BY~4.0.}}

\author{%
Thomas Goudemant\textsuperscript{1}, Benjamin Francesconi\textsuperscript{1}, Marjorie Bellizzi\textsuperscript{1}, Adrien Dorise\textsuperscript{1,2}\\[.4ex]
{\small\textit{\textsuperscript{1}IRT Saint Exup\'ery, France \quad \textsuperscript{2}CNES, France}}\\[.3ex]
{\small\texttt{thomas.goudemant@irt-saintexupery.com}}%
}

\date{Preprint --- accepted at OBPDC 2026, Barcelona, October 2026}

\begin{document}
\maketitle

\input{sections/00_abstract}

\input{sections/01_introduction}
\input{sections/02_related_work}
\input{sections/03_method}
\input{sections/04_experiments}
\input{sections/05_discussion}
\input{sections/06_conclusion}


{\footnotesize
\setlength{\bibsep}{0pt plus 0.2ex}
\linespread{0.96}\selectfont
\bibliography{references}
}

\end{document}

%% file: sections/00_abstract.tex
\begin{abstract}
Rapid assessment of building damage after natural disasters or conflict
events is essential to support emergency response and prioritise field
interventions. Earth Observation satellites can acquire relevant imagery
shortly after an event, but exploitation is limited by uplink and downlink
capacity and by ground-processing latency. We address this with a
bi-temporal building damage assessment pipeline built on a siamese detector
derived from YOLOX and designed to compress information at both ends of the
ground/space link. On the ground, pre-disaster reference images are encoded
into a compact latent space (compressed by up to a factor of~64) and
uplinked to the satellite. On board, this reference is compared with a fresh
post-disaster acquisition so that the downlink carries only actionable
object-level products, bounding boxes and damage classes, instead of full
scenes. This cuts the data exchanged in both directions and shortens the time
to an actionable result, while on xBD the strongly compressed reference still
preserves most of the detection performance.

Because on-board acquisitions suffer from residual pre/post co-registration
errors, we introduce a latent-space shift estimation and correction module
that regresses the global offset from the coarse feature level and realigns the
post-disaster features before fusion. It substantially improves robustness to
de-registration, especially under large shifts where fusion-only variants
collapse, while also raising nominal accuracy and remaining compatible with
the strongest compression. We finally port the pipeline to two embedded
targets, a Xilinx Versal VCK190 and an NVIDIA Jetson AGX~Orin, and report
deployment impact and hardware performance (latency, throughput, power
efficiency). We find that the core detector and its compression port cleanly to
both, but that the operators needed for long-range robustness survive only on
the Jetson GPU, which supports the full robust pipeline, whereas the Versal DPU
does not.
\end{abstract}

%% file: sections/01_introduction.tex
\section{Introduction}
\label{sec:intro}

Timely assessment of building damage after natural disasters or armed
conflicts is critical to support emergency response, allocate limited
resources and prioritise field interventions~\cite{gupta2019xbddatasetassessingbuilding}.
High-resolution (HR) Earth Observation (EO) satellites provide large-scale,
repeatable coverage of impacted regions, and deep learning has significantly
improved automated damage assessment from satellite
imagery~\cite{10645210,gupta2019xbddatasetassessingbuilding}. Most
high-performing methods formulate the task as a bi-temporal change detection
problem, jointly exploiting pre- and post-disaster imagery to disambiguate
newly destroyed structures from pre-existing ones~\cite{rs12101670,ZHENG2021112636}.
Such pipelines, however, remain largely ground-centric: they require full-image
downlink and offline processing before actionable products are
generated~\cite{rs15163963}, which delays decision-making and consumes
communication resources over areas of limited operational interest.

Advances in intelligent satellite systems and embedded AI accelerators enable
moving part of the processing closer to the sensor~\cite{9705087}, and on-board
object detection has been demonstrated for several EO
tasks~\cite{rs15163963,phisat2}. In this
context, on-board AI can be used as a \emph{data-reduction and prioritisation}
stage: rather than transmitting complete scenes, the satellite extracts and
downlinks only actionable information.

In our previous work~\cite{Goudemant_2026_CVPR}, we introduced a ground/on-board
architecture for bi-temporal building damage assessment that decouples pre- and
post-event processing: pre-disaster scenes are encoded on the ground into
compact latent representations, uplinked and stored on board, then compared with
newly acquired post-disaster images to produce object-level products. That study
established the benefit of siamese processing, latent-space compression and
shift augmentation, and used classical cross-attention as a reference mechanism
for robustness to residual co-registration errors. It did not, however, resolve
two open issues: (i) robustness to \emph{large} pre/post de-registration
degraded well before the model's effective receptive field was reached, and
(ii) the mechanisms giving robustness were not shown to be compatible with an
actual embedded target.

This paper extends that line of work along both axes.
\emph{First}, on the algorithmic side, we introduce a latent-space shift
estimation and correction module that explicitly regresses the global pre/post
offset from the coarse feature level and realigns the post-disaster features
before fusion. It substantially improves robustness to de-registration while
also raising nominal accuracy, and it remains compatible with aggressive latent
compression up to a factor of~$64$; detailed figures are reported in
Section~\ref{sec:experiments}.
\emph{Second}, on the deployment side, we port the on-board pipeline to two
representative embedded accelerators, a Xilinx Versal VCK190 and
an NVIDIA Jetson AGX~Orin, and report both the impact of
deployment on algorithmic performance and the hardware performance (latency,
throughput, power efficiency). We show that the core detection pipeline ports
cleanly, but that the operators required for long-range robustness (softmax
cross-attention and grid-sampling-based alignment) are either non-portable or
non-robust on the Versal DPU, whereas the Jetson GPU supports the full robust
pipeline.

\paragraph{Contributions.}
(i)~A latent-space shift estimation and correction module for robust
bi-temporal damage assessment under strong de-registration, compatible with
latent compression and designed for embedded (GPU) inference.
(ii)~An updated robustness analysis contrasting shift augmentation,
cross-attention, portable (DPU-compatible) attention, and the proposed
alignment module.
(iii)~An embedded deployment study on Versal VCK190 and Jetson AGX~Orin,
reporting algorithmic impact and hardware performance, and characterising which
robustness mechanisms survive each target.

%% file: sections/02_related_work.tex
\section{Related work}
\label{sec:related}

Building damage assessment from EO imagery is commonly cast as a bi-temporal
change detection problem, since post-only observation confuses damaged buildings
with background structures~\cite{rs12101670,ANDRESINI2023119123}; fully
convolutional siamese networks compare pre-/post-event images in a shared
representation space~\cite{DaudtSiamese}. Most benchmarks pose the task as dense
pixel-level segmentation~\cite{gupta2019xbddatasetassessingbuilding}, but
pixel-wise labelling can yield semantically inconsistent predictions at the
building scale when damage affects only part of a structure~\cite{ZHENG2021112636}.
Object-level formulations instead take the building as the unit of analysis,
localising each structure and assigning it a single damage state; this produces
more coherent, compact outputs~\cite{rs14071552} that are well suited to on-board
data reduction, where only actionable object-level products need be downlinked.
For robustness to residual co-registration, non-local mechanisms are attractive
because a feature can retrieve context from a displaced location: self-attention
does this~\cite{Vaswani2017,Mohammadian_2023}, but its spatial softmax is poorly
supported by fixed-function edge accelerators. Lightweight convolutional
attention such as CBAM~\cite{woo2018cbam} is hardware-friendly, but its spatial
component is a multiplicative gate with a fixed convolutional receptive field: it
re-weights features locally and cannot fetch information from a displaced
location, so it remains strictly local. On the deployment side, embedded AI in orbit has been demonstrated for
object and change detection on spaceborne
hardware~\cite{9705087,rs15163963,goudemant:hal-03881738,goudemant2024anomaly},
and system-level demonstrators have explored reactive ground--space
architectures that shorten the decision--action loop~\cite{francesconievent};
yet these efforts rarely combine bi-temporal reasoning, object-level modelling
and the constraints of embedded deployment~\cite{Dorise}. We refer the reader
to~\cite{Goudemant_2026_CVPR} for the full survey and adopt its object-level,
single-stage detection formulation; this paper focuses on the two new
contributions: an explicit latent-space alignment module and an embedded
portability study.

%% file: sections/03_method.tex
\section{Method}
\label{sec:method}

\subsection{Ground/on-board system concept}
\label{sec:concept}

We consider an operational EO scenario in which rapid building damage
assessment is required to support post-disaster response. Following a disaster
alert or the identification of an area of interest, the \emph{ground segment}
selects the relevant pre-disaster reference image(s), encodes them into compact
multi-scale latent representations, and uplinks only the latents over the
targeted area. The \emph{on-board segment} processes the newly acquired
post-disaster image and compares it with the stored pre-disaster latents to
produce compact object-level products: building bounding boxes and damage
classes, with the option to selectively downlink image crops over damaged
regions instead of full scenes.

\begin{figure}[t]
  \centering
  \includegraphics[width=\conceptwidth]{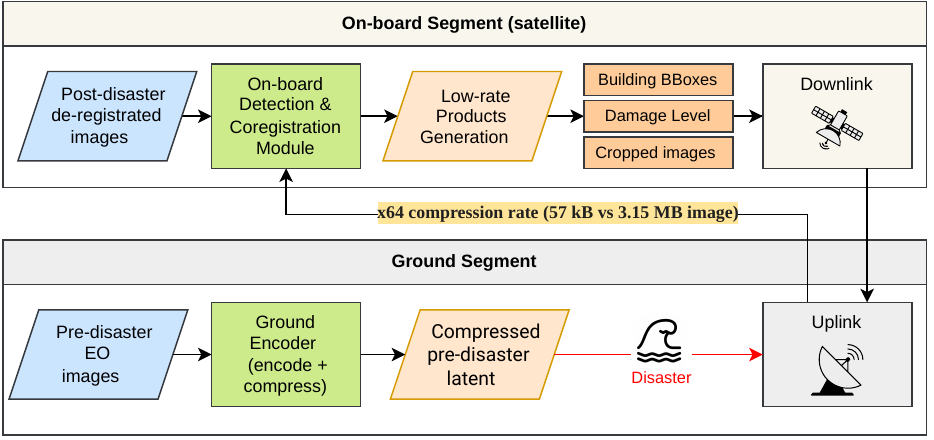}
  \caption{System concept. The \emph{ground segment} encodes pre-disaster
  references into a compressed latent ($\times64$) uplinked to the satellite; the
  \emph{on-board segment} detects and co-registers against the fresh
  post-disaster image and downlinks only low-rate products (bounding boxes,
  damage level, optional crops over damaged regions) instead of full scenes.}
  \label{fig:concept}
\end{figure}

This separation (Fig.~\ref{fig:concept}) reduces the uplink volume (compact
latents instead of full-resolution pre-disaster images) and the downlink volume
(damage products and selected crops instead of full post-disaster scenes). It
also introduces the constraints that drive our design: uplink bandwidth and
on-board memory forbid transmitting complete pre-disaster scenes, and limited
on-board geolocation accuracy induces residual pre/post misregistration. The
detailed network is shown in Fig.~\ref{fig:align}.

\subsection{Base detection architecture}
\label{sec:base}

The base detector and the two baseline robustness mechanisms below (shift
augmentation and attention) are those of~\cite{Goudemant_2026_CVPR}, where they
are detailed; we only summarise them here.
The backbone is YOLOX-S~\cite{ge2021yolox}, whose CSPDarknet produces three
feature maps (D3, D4, D5) at strides $8$, $16$ and $32$. These levels differ in
nature: D3, at stride~$8$ and the finest spatial resolution, stays close to the
input image and retains fine geometric detail, which makes it the level on which
spatial registration can still be performed; D5, at stride~$32$, is spatially
coarse but semantically rich, encoding object-level representations. YOLOX's
Path Aggregation FPN (PAFPN)~\cite{lin2017fpn} then fuses these maps with both a
top-down and a bottom-up path, propagating D5's semantic context into the finer,
better-localised D3/D4 levels and, in turn, their spatial detail back up.

We start from a classic YOLOX and simply feed it the pre- and post-disaster
images stacked along the channel axis (a $6$-channel input); we refer to this
baseline as \emph{early fusion}, since the two images are mixed at the very
input. We then evolve it into a \emph{siamese} architecture: a single shared
encoder processes the pre- and the post-disaster image independently, and the two
feature sets are concatenated, together with their element-wise difference,
before the FPN. This decomposition lets the network reason separately about each
image (e.g.\ distinguishing building edges from ground) while learning a common
representation of both, which improves detection over \emph{early fusion}. Crucially,
sharing one encoder maps directly onto the ground/on-board split: the same
encoder is used on the ground to encode the pre-disaster reference and on board
to encode the newly acquired post-disaster image, so pre-disaster features are
computed on the ground and reused on board.

\paragraph{Compressed pre-disaster latents.}
The raw multi-scale pre-disaster features exceed the input image size
($\approx3.7$\,MB vs.\ $3.15$\,MB at $1024^2$), so a lightweight $1\times1$
convolution reduces the channel count by a factor $r$ before uplink, with a
symmetric projection restoring dimensionality on board.
Tab.~\ref{tab:latent} reports the resulting int8 footprint: $r{=}64$ shrinks the
latent to $57$\,kB, two orders of magnitude below the image, while $r{=}128$
leaves a single latent channel at the finest level (D3), which we show is
insufficient for the alignment module (Sec.~\ref{sec:exp-robust}).

\begin{table}[htbp]
\centering
\caption{Size of the pre-disaster latent representation under channel
compression, for a $1024\times1024\times3$ input, reported as an equivalent
int8 footprint. Spatial resolutions: $128^2$ (D3), $64^2$ (D4), $32^2$ (D5).}
\label{tab:latent}
\resizebox{\tabwidthnarrow}{!}{%
\begin{tabular}{lcccc}
\toprule
\textbf{Configuration} & \textbf{D3} & \textbf{D4} & \textbf{D5} & \textbf{Size} \\
\midrule
Input image ($1024^2{\times}3$) & -- & -- & -- & 3.15\,MB \\
Latent (no comp.)               & 128 & 256 & 512 & 3.67\,MB \\
Latent (comp.\ $r{=}8$)         & 16  & 32  & 64  & 0.46\,MB \\
Latent (comp.\ $r{=}64$)        & 2   & 4   & 8   & 57\,kB \\
Latent (comp.\ $r{=}128$)       & 1   & 2   & 4   & 29\,kB \\
\bottomrule
\end{tabular}}
\end{table}

\subsection{Robustness to pre/post de-registration}
\label{sec:robust-methods}

Residual spatial misregistration between the pre- and post-disaster images is
unavoidable on board, due to acquisition-geometry differences and limited
on-board registration. Under such shifts, strictly local feature comparison
fails to retrieve the relevant pre-disaster information at the correct location,
which mostly degrades damage classification (localisation is more stable,
as boxes are predicted in the pre-disaster reference
frame)~\cite{Goudemant_2026_CVPR}. We consider three
families of mechanisms, of increasing effectiveness.

\paragraph{(a) Shift augmentation and (b) attention.}
Shift augmentation applies, during training, a random displacement (offsets in
$[0,150]$\,px) to the post-disaster image only, with ground truth kept in the
pre-disaster frame; it is used in all variants unless stated. Cross-attention
lets each pre-disaster location attend over the whole post-disaster map and
retrieve possibly shifted features, using the query--key--value formulation
of~\cite{Vaswani2017} on the coarse levels (D4--D5); its
$\mathrm{HW}\times\mathrm{HW}$ softmax, however, is not supported by the Versal
DPU (Sec.~\ref{sec:portability}). We therefore additionally consider
\emph{portable} attention gates built only from DPU-compatible operators, the
most representative being a \emph{spatial-difference} gate adapted from the
spatial branch of CBAM~\cite{woo2018cbam}: a depthwise $7\times7$ convolution
over the signed pre$-$post difference, followed by a pointwise convolution and a
sigmoid gate. Being purely convolutional, it has a strictly local receptive
field.

\begin{figure*}[htbp]
  \centering
  \includegraphics[width=\archwidth]{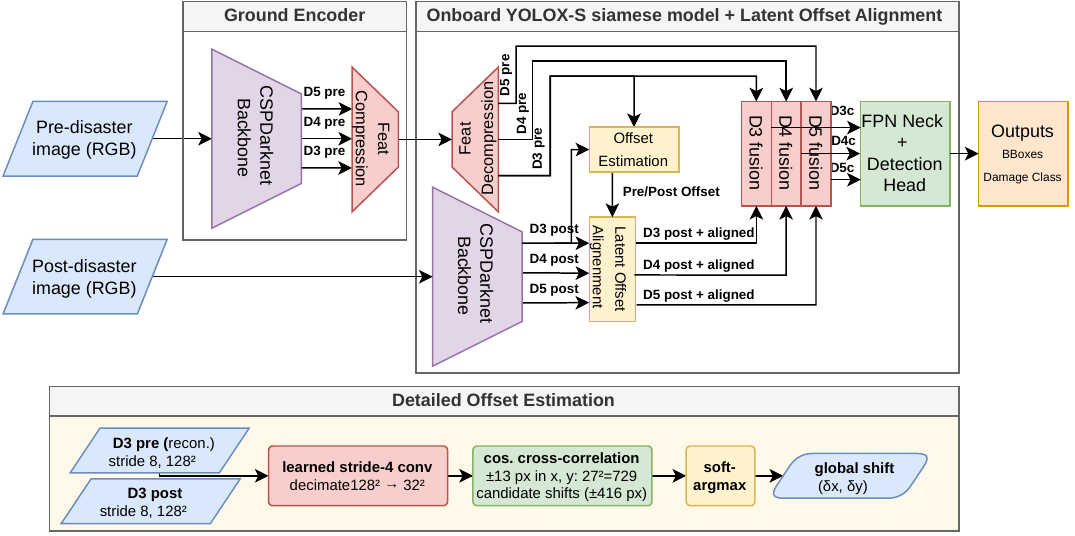}
  \caption{Proposed ground/on-board architecture (details in
  Sec.~\ref{sec:robust-methods}). Compressed pre-disaster latents are uplinked
  and expanded on board; the offset-estimation module regresses a global shift
  $(\delta_x,\delta_y)$ from the D3 latent, which the latent offset alignment
  applies to realign the post D3/D4/D5 features before multi-scale fusion.}
  \label{fig:align}
\end{figure*}

\paragraph{(c) Latent-space shift estimation and correction (proposed).}
We estimate the global pre/post offset explicitly and use it to realign the
post-disaster features before fusion (Fig.~\ref{fig:align}). An \emph{offset head}
predicts one global displacement $(\delta_x,\delta_y)$ per image by
cross-correlating the pre and post features. It works on D3, the finest and
least-compressed level, whose geometric detail is best for matching; a stride-$4$
convolution first subsamples it to $32\times32$ to keep the search cheap. On a
grid of $(2R{+}1)^2$ candidate offsets, we cosine-correlate the pre and shifted
post embeddings, and a soft-argmax over the grid gives a sub-pixel offset. We use
$R{=}13$, covering shifts of up to $416$\,px. A coarse grid is sufficient in
practice: the correction only needs to bring the residual shift back within the
perceptive field, where detection stays robust ($\approx20$\,px,
Fig.~\ref{fig:robustness}), and the soft-argmax already interpolates below one
cell. The offset then shifts the post-disaster D3/D4/D5
features toward the pre features by bilinear resampling, each level scaled by its
stride. The whole module is differentiable, so it is trained end-to-end by the
detection loss plus an auxiliary loss that regresses the known training shift.

The module stays exportable on embedded accelerators: the offset search is a
single vectorised correlation and the resampling reduces to one grid-sampling
operation. Grid-sampling is a native TensorRT operator, so the alignment is
portable on the Jetson GPU but not on the Versal DPU
(Sec.~\ref{sec:portability}), which it therefore targets.

\subsection{Embedded targets and portability}
\label{sec:portability}

We deploy the on-board branch on two representative accelerators.
The \textbf{Xilinx Versal VCK190}, whose adaptive-compute
architecture~\cite{gaide2019versal} is a candidate for next-generation edge
computing in space~\cite{perryman2023versal}, carries a
\texttt{DPUCVDX8G\_ISA3\_C32B6} DPU (batch~6) programmed through Vitis~AI with
int8 quantisation and a fixed-function operator set. The \textbf{NVIDIA Jetson
AGX~Orin}~\cite{barnell2022jetson} is a general-purpose edge GPU programmed
through TensorRT, capped here to a $30$\,W power mode.

The DPU imposes a hard portability constraint: the on-board model must compile to
a \emph{single} DPU subgraph, otherwise the runtime aborts. The siamese detector
and its latent-compressed variants meet this constraint and port cleanly, as does
the portable attention gate (the spatial-difference gate). Classical
cross-attention and the alignment module, by contrast, are not portable to the
DPU. In short, the DPU supports the core bi-temporal detector, compression and
the portable attention gate, but not the operators providing long-range
robustness through cross-attention or explicit alignment, which the Jetson GPU
does support.

%% file: sections/04_experiments.tex
\section{Experiments and results}
\label{sec:experiments}

\subsection{Dataset and protocol}
\label{sec:exp-setup}

We conduct all experiments on the xBD
dataset~\cite{gupta2019xbddatasetassessingbuilding} (xView2 Challenge), built
from very-high-resolution optical imagery of the Maxar/DigitalGlobe Open Data
program. The images are orthorectified RGB products, provided as coarsely
registered pre-/post-event pairs of size $1024\times1024$ at a ground sampling
distance below $0.8$\,m/px. Buildings are
annotated with four damage levels (\emph{no damage}, \emph{minor},
\emph{major}, \emph{destroyed}). We reformulate the segmentation annotations as
object detection by converting polygons to axis-aligned boxes, and use the
standard split (excluding Tier~3): $4{,}665$ pairs, $60/20/20\%$ train/val/test.
All variants use a YOLOX-S backbone and identical optimisation settings.
Detection is evaluated with precision, recall, F1 and mAP@0.5. Robustness is
measured by applying synthetic shifts at test time to the post-disaster image
only, with ground truth kept in the pre-disaster frame and all metrics computed
on the common spatial support induced by the maximum displacement, so that every
shift is scored on an identical image and annotation subset.

\subsection{Nominal detection performance}
\label{sec:exp-nominal}

Tab.~\ref{tab:nominal} reports performance without test-time shift. Siamese
processing improves over \emph{early fusion}, confirming the benefit of separating pre-
and post-disaster feature extraction before comparison. Cross-attention and the
portable spatial-difference gate leave nominal accuracy essentially unchanged.
The proposed shift estimation and correction module gives the best nominal
mAP@0.5 ($57.8\%$), and combining it with latent compression $r{=}64$ costs only
$\approx1$\,pp of mAP while shrinking the uplinked latent by two orders of
magnitude (Tab.~\ref{tab:latent}). This confirms that the pre-disaster
information required for the task can be represented with a much smaller volume
than the original image.

\begin{table}[htbp]
\centering
\caption{Nominal detection performance on xBD (no test-time shift), \%.
All variants use shift augmentation unless stated.}
\label{tab:nominal}
\resizebox{\tabwidth}{!}{%
\begin{tabular}{lcccc}
\toprule
\textbf{Configuration} & \textbf{P} & \textbf{R} & \textbf{F1} & \textbf{mAP@0.5} \\
\midrule
Early Fusion (baseline)          & 58.3 & 55.0 & 56.3 & 55.1 \\
Siamese (no shift-aug)           & 59.0 & 58.0 & 58.2 & 56.4 \\
Siamese                          & 58.5 & 57.4 & 57.8 & 56.6 \\
Siamese + Cross-Attention        & 61.1 & 56.0 & 58.4 & 56.5 \\
Siamese + Portable Attn (Sp.-Diff) & 58.9 & 57.3 & 57.9 & 56.7 \\
Siamese + Shift Est.\ \& Corr.    & 61.5 & 57.3 & 59.2 & \textbf{57.8} \\
\;\;+ Latent Compression $\times64$ & 61.3 & 55.8 & 58.3 & 56.7 \\
\bottomrule
\end{tabular}}
\end{table}

\subsection{Robustness to de-registration}
\label{sec:exp-robust}

Fig.~\ref{fig:robustness} reports mAP@0.5 as a function of the applied
post-disaster shift. Without shift augmentation (the \emph{early-fusion} and
\emph{siamese-no-aug} baselines) performance collapses within a few tens of pixels,
i.e.\ typical building extents. Shift augmentation alone slows the degradation but
its mAP@0.5 still falls to $26.6\%$ at a $400$-pixel shift (from $56.6\%$
nominal). Cross-attention adds genuine long-range robustness ($43.0\%$ at
$400$\,px) but does not port to the DPU (it runs on the Jetson GPU only), while
the portable spatial-difference gate behaves like the plain siamese model
($29.2\%$): its local receptive field cannot retrieve distant features, so it does
not deliver the robustness of true attention. This is a negative result we make
explicit, since it is exactly this gap that motivates an explicit
alignment.

The proposed shift estimation and correction module is the clear best variant: it
keeps mAP essentially flat across the whole shift range, holding $49.4\%$ at
$400$\,px (from $57.8\%$ nominal), against $26.6\%$ for shift augmentation and
$43.0\%$ for cross-attention, while also improving nominal accuracy. Notably, mAP
stays flat well beyond the $150$\,px training range of the shift augmentation
(vertical line in Fig.~\ref{fig:robustness}): the model extrapolates to
shifts larger than any seen during training, because the offset head estimates
the displacement geometrically rather than memorising trained offsets. Adding
latent compression $\times64$ preserves this behaviour almost exactly ($47.0\%$
at $400$\,px),
so aggressive data reduction and strong de-registration tolerance are compatible.

\ifsidebyside
\begin{figure*}[htbp]
  \centering
  \begin{minipage}[t]{0.49\textwidth}
    \centering
    \includegraphics[width=\linewidth]{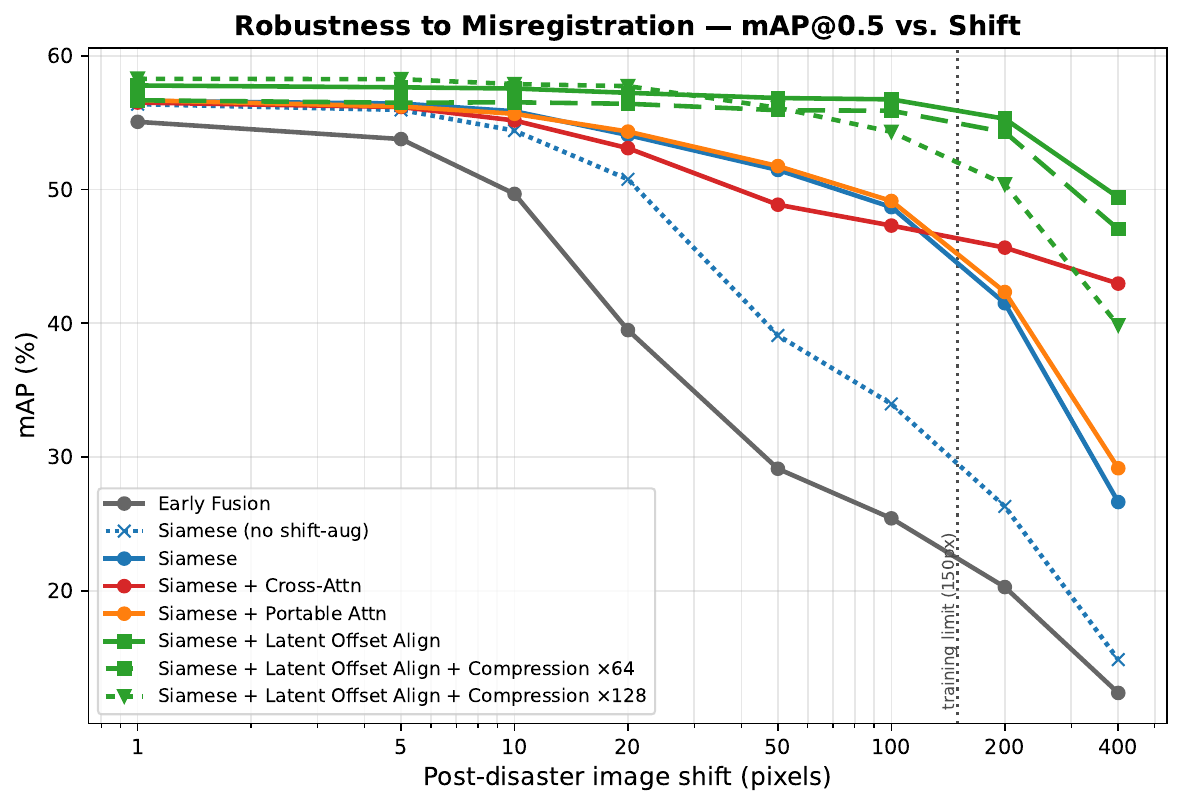}
    \caption{Robustness to pre/post misregistration on xBD: mAP@0.5 vs.\ applied
    post-disaster shift (log-$x$), averaged over shift directions. The vertical
    line marks the $150$\,px training range of the shift augmentation; the
    proposed module (green) stays flat beyond it, whereas shift-augmentation and
    attention baselines degrade.}
    \label{fig:robustness}
  \end{minipage}\hfill
  \begin{minipage}[t]{0.49\textwidth}
    \centering
    \includegraphics[width=\linewidth]{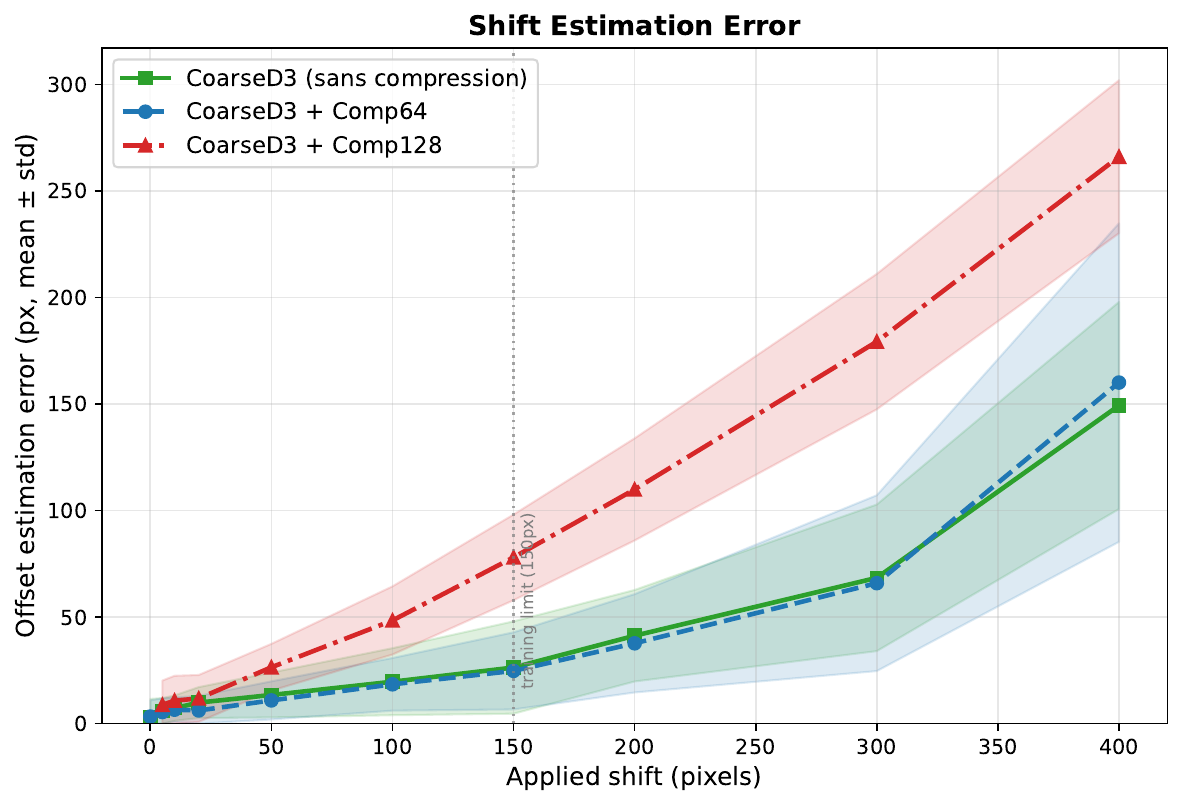}
    \caption{Offset estimation error (mean~$\pm$~std) vs.\ applied shift.
    Compression $\times64$ matches the uncompressed estimator; $\times128$
    (a single D3 latent channel) roughly doubles the error and breaks alignment.
    The rise beyond $\sim150$\,px reflects extrapolation outside the training
    range.}
    \label{fig:offset}
  \end{minipage}
\end{figure*}
\else
\begin{figure}[htbp]
  \centering
  \includegraphics[width=0.99\linewidth]{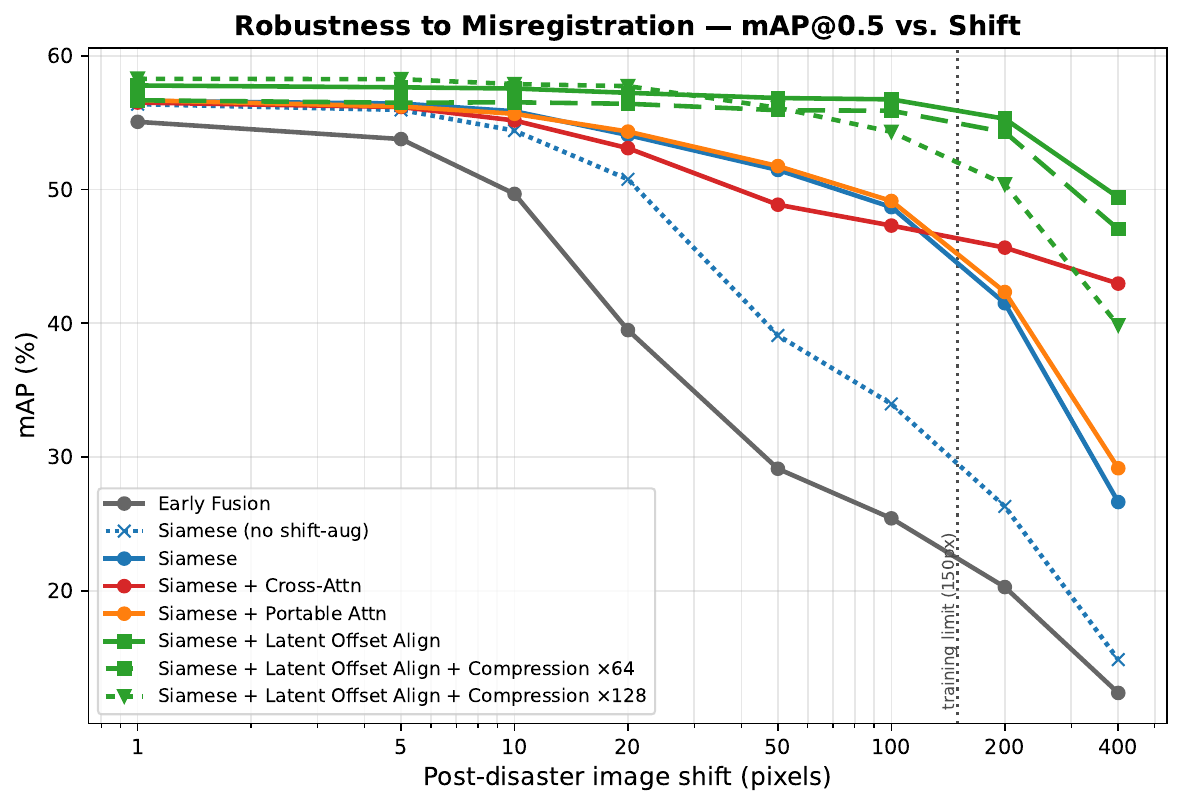}
  \caption{Robustness to pre/post misregistration on xBD: mAP@0.5 vs.\ applied
  post-disaster shift (log-$x$), averaged over shift directions. The vertical
  line marks the $150$\,px training range of the shift augmentation; the proposed
  module (green) stays flat beyond it, whereas shift-augmentation and attention
  baselines degrade.}
  \label{fig:robustness}
\end{figure}
\fi

\paragraph{Offset estimation quality and the $r{=}128$ limit.}
Fig.~\ref{fig:offset} reports the module's offset estimation error. It is low
and grows smoothly within the useful range, and compression $\times64$ tracks
the uncompressed estimator almost exactly, confirming that the offset can be
recovered directly from the compact latent. Pushing to $\times128$, however,
leaves a single latent channel at D3 (Tab.~\ref{tab:latent}): the correlation signal becomes too weak
and the estimation error roughly doubles. The alignment can then no longer lock
onto the shift, and this propagates to the end task: the post features are
mis-warped, so damage classification degrades and the overall detection
mAP@0.5 drops well below the $\times64$ configuration under shift, undoing the
robustness gain. $r{=}128$ removes the information the alignment module needs, so
$r{=}64$ is the aggressive-compression limit at which the robust pipeline still
works.

\ifsidebyside\else
\begin{figure}[htbp]
  \centering
  \includegraphics[width=0.95\linewidth]{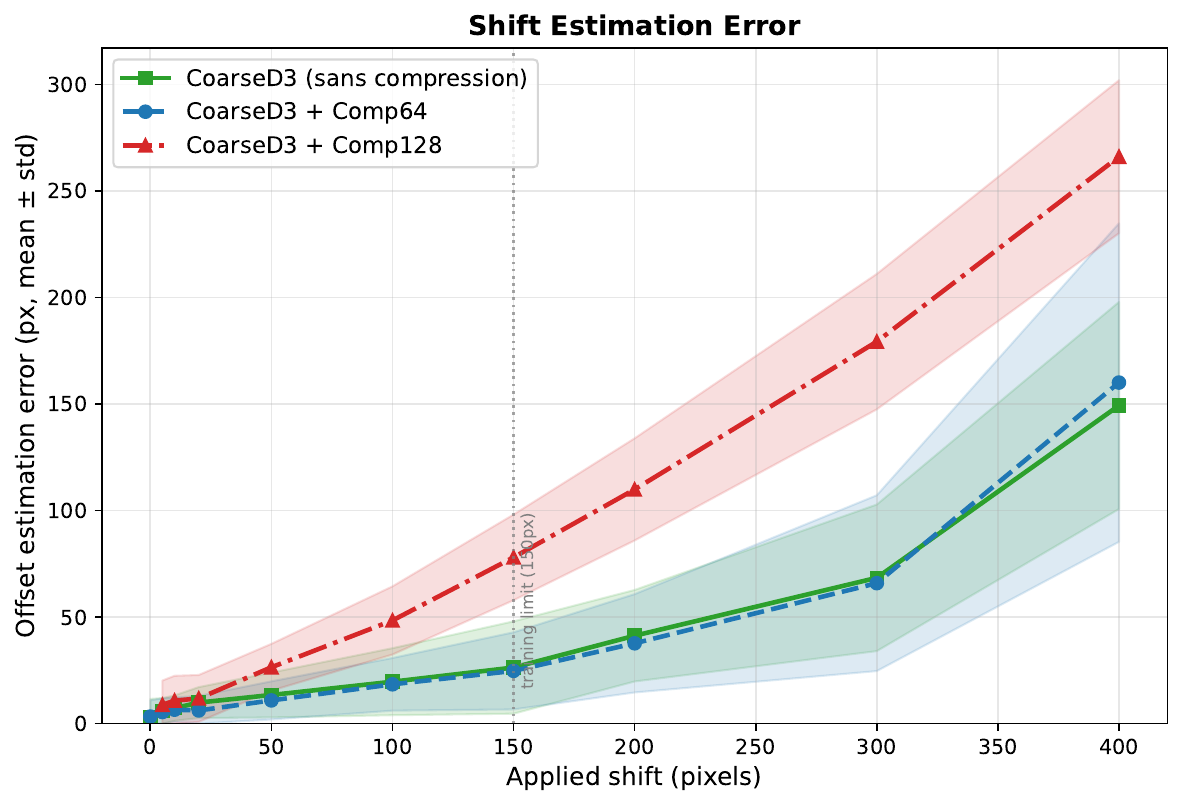}
  \caption{Offset estimation error (mean~$\pm$~std) vs.\ applied shift.
  Compression $\times64$ matches the uncompressed estimator; $\times128$
  (a single D3 latent channel) roughly doubles the error and breaks alignment.
  The rise beyond $\sim150$\,px reflects extrapolation outside the training
  range.}
  \label{fig:offset}
\end{figure}
\fi

\subsection{Embedded deployment: algorithmic impact}
\label{sec:exp-deploy-algo}

Tab.~\ref{tab:quant} reports the effect of porting on mAP@0.5. Quantisation is
near-neutral for all compressed/portable models on both targets: on the Jetson,
the plain siamese detector loses only $0.31$\,pp in int8; on the Versal, the
$\times128$-compressed detector and the portable spatial-difference proxy stay
within $\pm1.3$\,pp of their float32 references. The exception is the full
alignment model (with compression $\times64$) on the Jetson, which drops
$3.24$\,pp \emph{already in FP16}.

We attribute this to a porting bug in the alignment path, not to numerical
precision. Two arguments support this. First, FP16 is otherwise near-lossless on
this target (all other models lose $<0.4$\,pp), so a $3$\,pp drop appearing in
FP16, before any int8 rounding, cannot be a quantisation effect. Second,
the drop is localised to the one component that distinguishes this model: the
grid-sampling warp of the alignment module. The operator does export and run (the
engine builds and produces detections), but its numerical output diverges from
the reference PyTorch model. The likely cause is a mismatch in the resampling
convention between the two implementations (corner alignment, the
normalisation of sampling coordinates, or the border-padding mode), any of
which shifts the warped features by a fraction of a cell and thus corrupts the
pre/post comparison that damage classification relies on. This is an
implementation-level porting issue, still open at the time of writing, and must
be resolved before this model is trusted in production on the Jetson; it does not
affect the other reported models, none of which use grid sampling.

\begin{table}[htbp]
\centering
\caption{Impact of quantisation/deployment on porting, per model
(mAP@0.5, \%). Rows compare each model's float32 reference against its
on-target quantised version, not models across platforms.}
\label{tab:quant}
\setlength{\tabcolsep}{4pt}
\resizebox{\tabwidth}{!}{%
\begin{tabular}{llccc}
\toprule
\textbf{Platform} & \textbf{Model} & \textbf{float32} & \textbf{Quant.} & $\boldsymbol{\Delta}$ \\
\midrule
Jetson (TensorRT) & Siamese                 & 57.90 & 57.59 (int8) & $-0.31$ \\
Jetson (TensorRT) & S.E.C.\ + C$\times$64    & 57.63 & 54.39 (FP16) & $-3.24^\ast$ \\
VCK190 (Vitis AI) & Siamese + C$\times$128   & 52.66 & 53.98 (int8) & $+1.32$ \\
VCK190 (Vitis AI) & Portable Sp.-Diff        & 58.85 & 58.03 (int8) & $-0.81$ \\
\bottomrule
\end{tabular}}

\vspace{2pt}
{\scriptsize $^\ast$~open porting issue on the grid-sampling alignment path,
not a quantisation effect.}
\end{table}

\subsection{Embedded deployment: hardware performance}
\label{sec:exp-deploy-hw}

\begin{figure*}[t]
  \centering
  \includegraphics[width=\hwjetsonwidth]{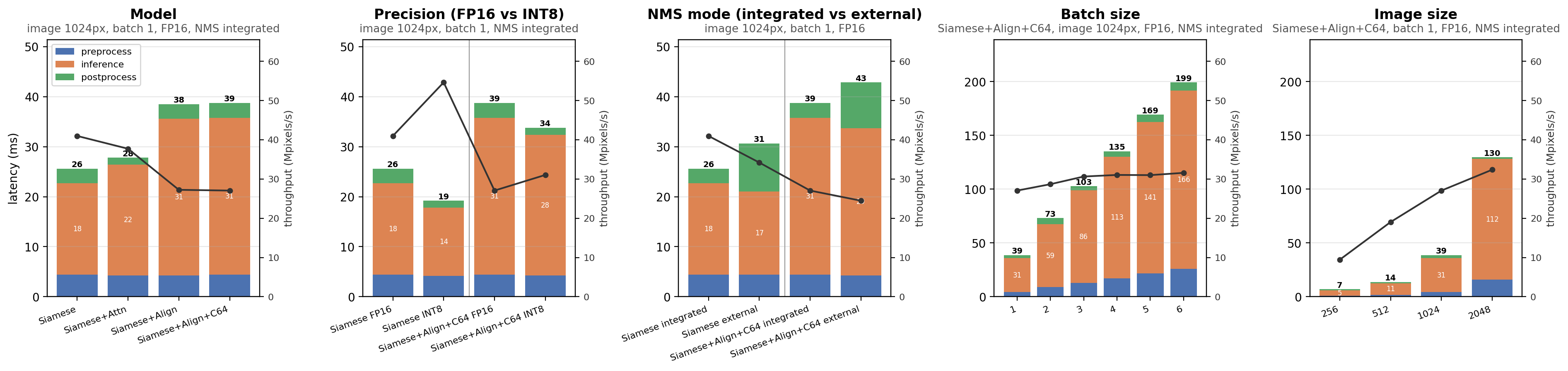}\\[3pt]
  \includegraphics[width=\hwjetsonwidth]{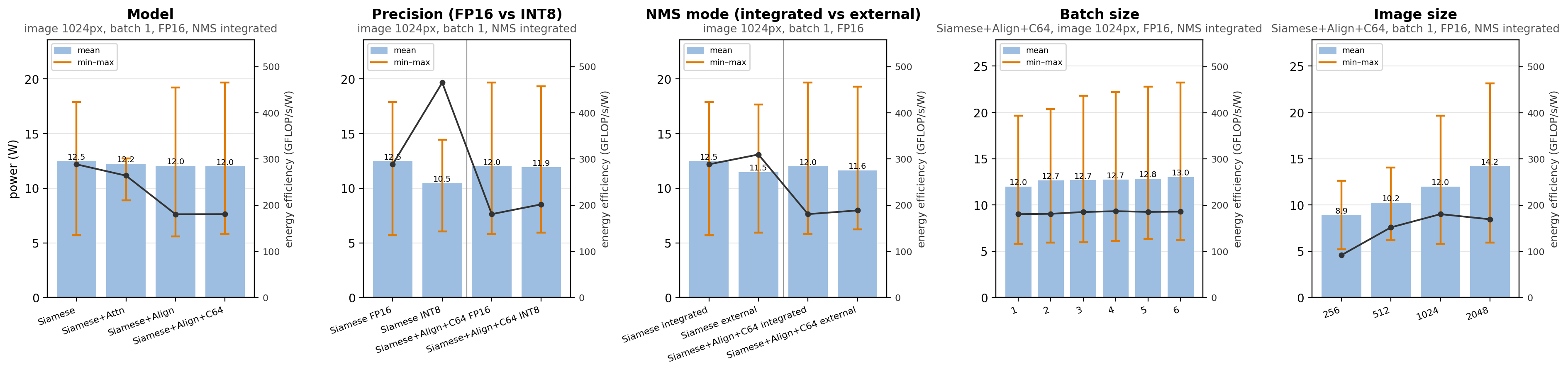}
  \caption{Jetson AGX~Orin sweep. Top: latency (stacked bars: preprocess /
  inference / postprocess, left axis) and throughput (line, right axis). Bottom:
  power (mean bar with a min--max whisker, left axis) and energy efficiency
  (line, right axis). Five axes: model, precision, NMS mode, batch size, image
  size.}
  \label{fig:hw-jetson}
\end{figure*}

\begin{figure*}[t]
  \centering
  \includegraphics[width=0.225\linewidth]{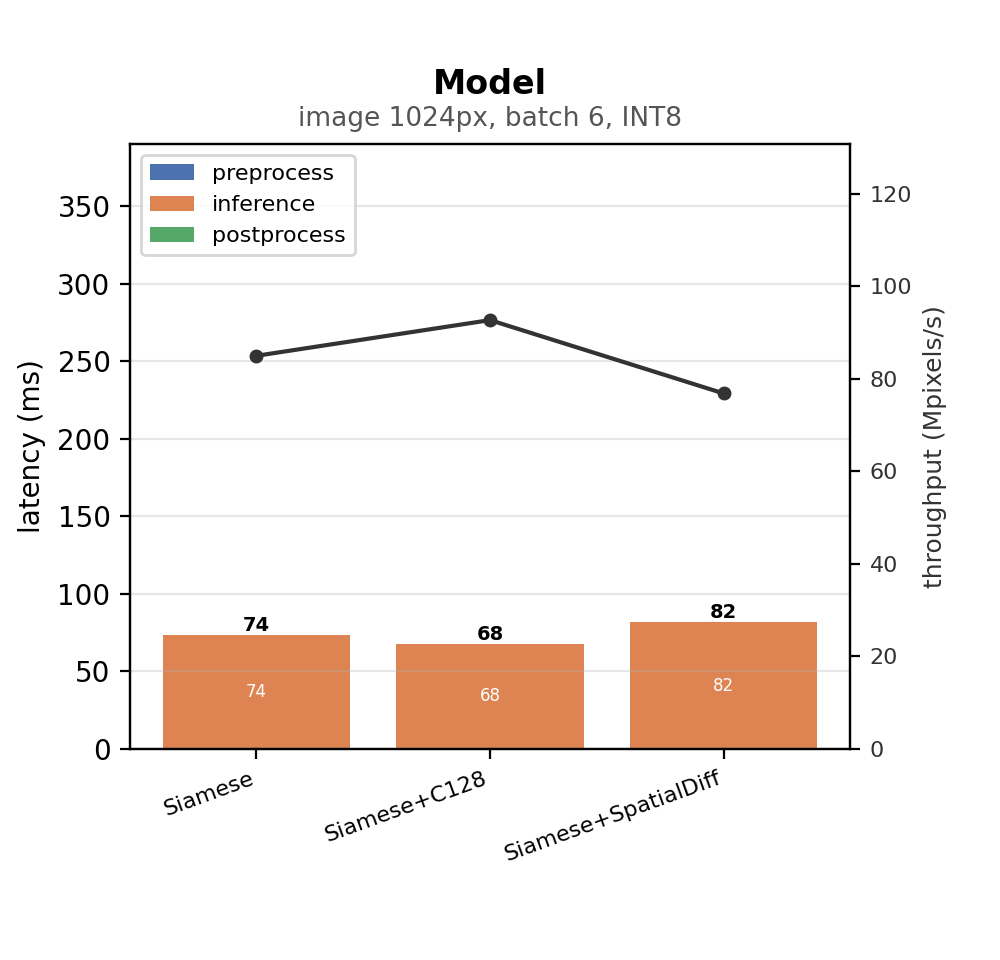}\hfill
  \includegraphics[width=0.225\linewidth]{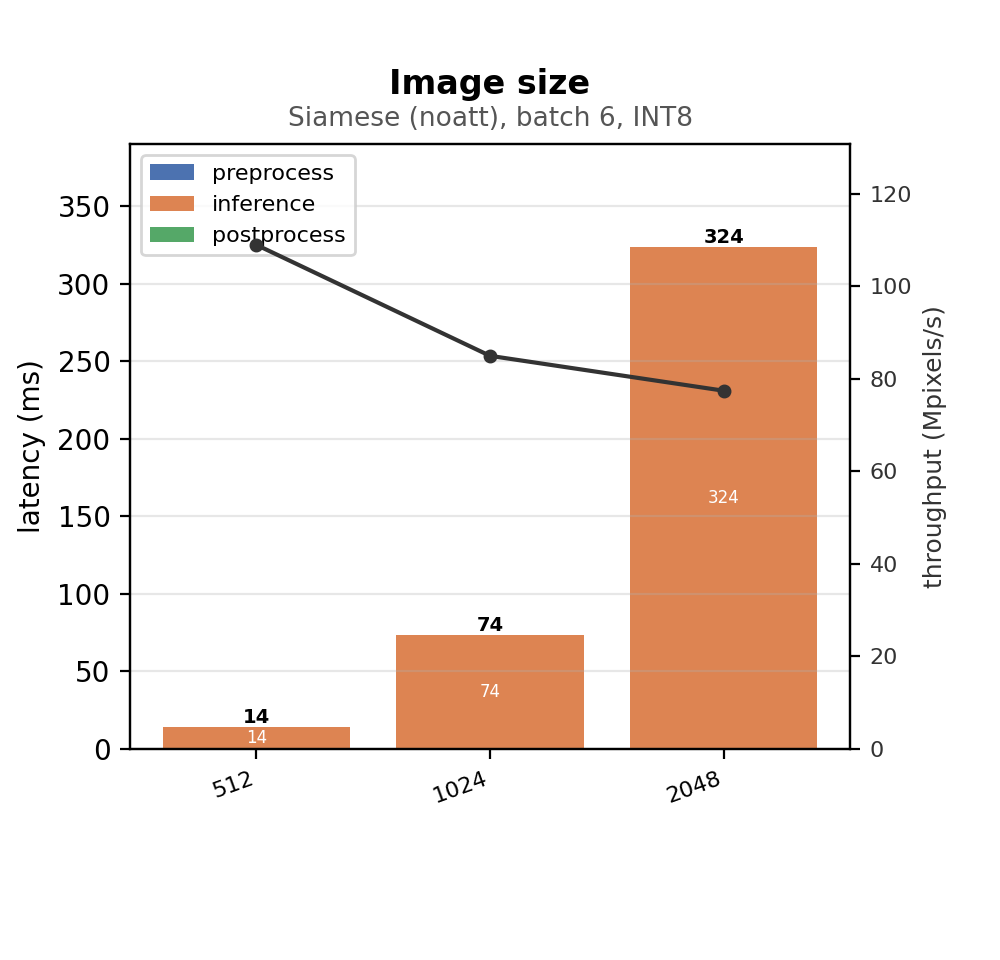}\hfill
  \includegraphics[width=0.225\linewidth]{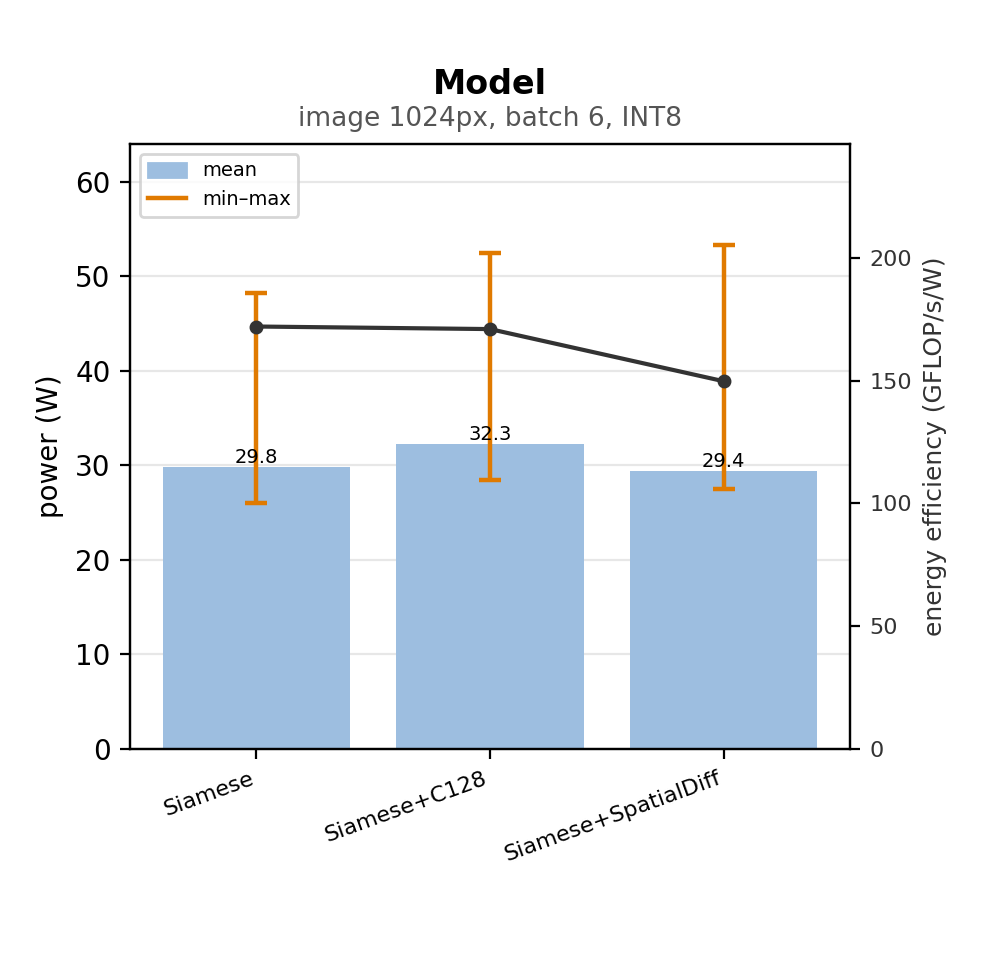}\hfill
  \includegraphics[width=0.225\linewidth]{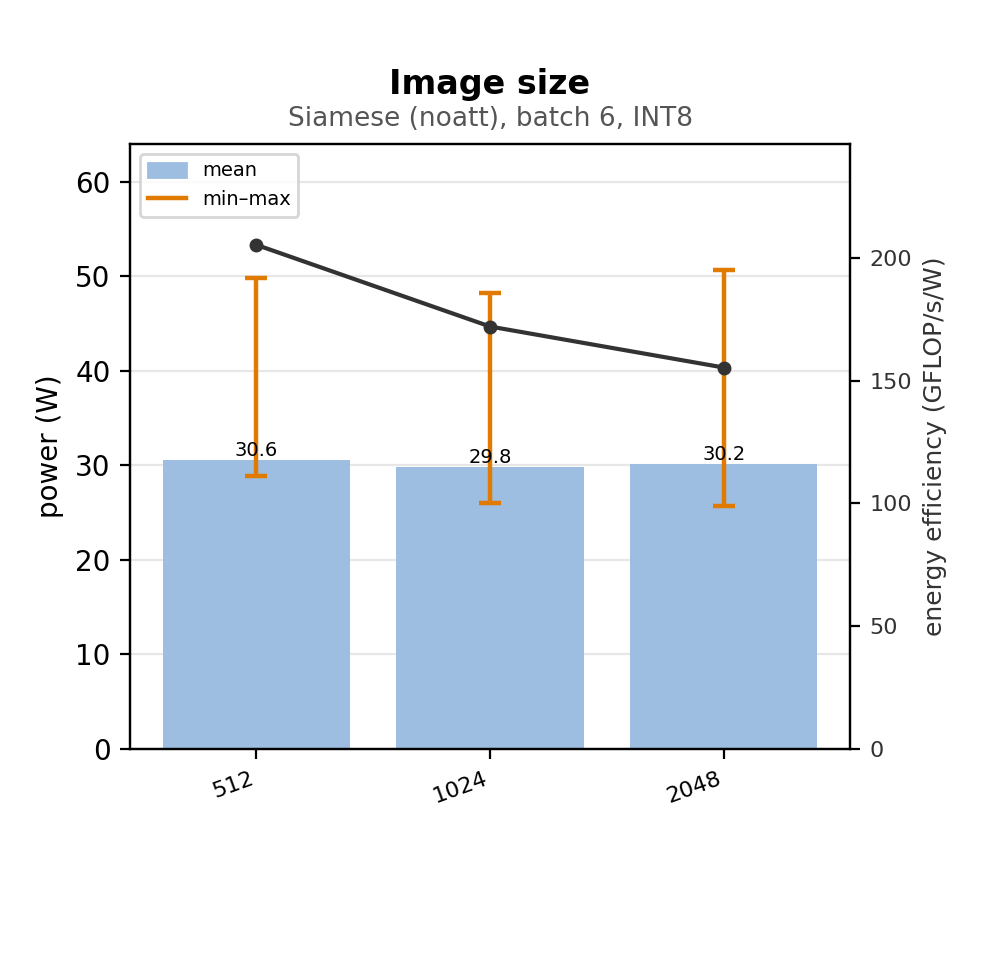}
  \caption{VCK190 sweep (int8, batch~6), two axes (model, image size). From left
  to right: latency/throughput by model, latency/throughput by image size,
  power/efficiency by model, power/efficiency by image size. The DPU batch size
  is fixed and the NMS/precision knobs of the Jetson path do not apply.}
  \label{fig:hw-vck190}
\end{figure*}

We measure end-to-end latency (broken down into preprocess / inference /
postprocess), throughput and power on both targets. On the Jetson we sweep five
axes: model, precision (FP16/int8), NMS mode, batch size
and image size (Fig.~\ref{fig:hw-jetson}); NMS runs either as a CPU
post-processing step (\emph{external}) or fused into the engine as a GPU TensorRT
plugin (\emph{integrated}, faster); on the VCK190 we can only sweep model and
image size (Fig.~\ref{fig:hw-vck190}): the other axes are fixed in the DPU image
(changing the batch size requires rebuilding it, too costly to sweep), and NMS
runs on the CPU, unsupported by the DPU.
Unless stated, the reference point is $1024\times1024$, integrated NMS, Jetson at
batch~$1$/FP16/$30$\,W mode and VCK190 at batch~$6$/int8. Power is read from each
board's on-board monitors: the Jetson's three GPU/CPU/SYS rails (INA3221) and the
VCK190's $17$~INA226 rails. Throughput is reported in Mpixels/s, which is
comparable across image sizes and batch settings.

\paragraph{Per-model behaviour.}
On the Jetson, the plain siamese detector runs at $25.6$\,ms / $41.0$\,Mpx/s /
$288$\,GFLOP/s/W; adding cross-attention is almost free ($27.8$\,ms), whereas the
alignment module is $\approx40\%$ slower ($38.5$--$38.8$\,ms) because of the
multi-level grid-sampling warp, not because of attention. Latent
compression $\times64$ brings \emph{no} latency gain over the uncompressed
alignment model ($38.8$ vs.\ $38.5$\,ms): it shrinks the tensor exchanged between
ground and edge, which matters for the uplink, but does not change the on-board
compute that dominates latency. On the VCK190, compression $\times128$ is instead
the fastest model ($67.9$\,ms / $92.7$\,Mpx/s), $8\%$ faster than the plain
detector, because DPU compute is bound by the feature volume and compression
lightens it directly; the portable spatial-difference proxy is the slowest
($81.9$\,ms), the cost of its extra spatial gate.

\paragraph{Precision, NMS and image/batch sweeps.}
The Jetson sweep (Fig.~\ref{fig:hw-jetson}) isolates several system knobs.
\emph{Precision:} switching FP16 to int8 cuts latency by $\approx13\%$ and lifts
efficiency markedly (up to $465$\,GFLOP/s/W on the plain detector) at
near-constant power. \emph{NMS mode:} integrated NMS is consistently $3$--$5$\,ms
cheaper in postprocess than external NMS, since it avoids the GPU-to-CPU
round-trip; we thus use it for all hardware measurements, external NMS being
needed only for uncapped-detection mAP evaluation. \emph{Batch size:} throughput
rises from batch~$1$ to $\approx3$ ($+13\%$) and then plateaus, as the GPU
already saturates at $1024^2$; power stays nearly flat. \emph{Image size:}
throughput in Mpixels/s grows with resolution ($9.4\to32.4$\,Mpx/s from $256$ to
$2048$) as the fixed per-pixel cost is better amortised, while frame rate falls
and power rises with load. On the VCK190, $512$\,px is the most efficient
operating point ($3.6$ vs.\ $2.6$\,Mpx/s/W at $1024$/$2048$).

\paragraph{Cross-platform summary.}
Tab.~\ref{tab:hw} collects the best portable model per target plus the common
\emph{noatt} reference. On the plain detector (the only fair cross-platform
comparison), the VCK190 reaches $\approx2\times$ the throughput of the Jetson
($85$ vs.\ $41$\,Mpx/s) but draws $\approx2.4\times$ the power
($29.8$ vs.\ $12.5$\,W), so energy efficiency favours the Jetson
($288$ vs.\ $172$\,GFLOP/s/W), consistent with a general-purpose GPU tuned
for FP16/int8 versus an int8-only DPU on a development board. Each platform's
best model behaves differently. On the VCK190 compression is essentially free:
the compressed best model matches the plain detector in efficiency ($171$ vs.\
$172$\,GFLOP/s/W) and is even marginally faster. On the Jetson, enabling the full
robust pipeline (S.E.C.\ alignment) is what costs: efficiency drops from $288$ to
$181$\,GFLOP/s/W and latency rises $25.6\to38.8$\,ms ($\approx40\%$), bringing the
Jetson down to roughly the Versal's efficiency, in exchange for the long-range
robustness the DPU cannot provide.

\begin{table}[htbp]
\centering
\caption{Hardware performance at $1024^2$. Only \emph{noatt} (Siamese) is
compared across platforms; per-platform best models use different compression and
are not cross-compared. Jetson: FP16, batch~1, $30$\,W; VCK190: int8, batch~6.}
\label{tab:hw}
\setlength{\tabcolsep}{3pt}
\resizebox{\tabwidth}{!}{%
\begin{tabular}{llccccc}
\toprule
\textbf{Platform} & \textbf{Model} & \textbf{GFLOPs} & \textbf{Lat.} & \textbf{Thr.} & \textbf{Power} & \textbf{Eff.} \\
 & & (/img) & (ms) & (Mpx/s) & (W) & (GF/s/W) \\
\midrule
Jetson Orin & Siamese                 & 69.8 & 25.6 & 41.0 & 12.5 & 288 \\
VCK190      & Siamese                 & 69.8 & 74.1 & 85.0 & 29.8 & 172 \\
Jetson Orin & S.E.C.\ + C$\times$64\, (best) & 70.4 & 38.8 & 27.1 & 12.0 & 181 \\
VCK190      & Siamese + C$\times$128\, (best) & 69.9 & 67.9 & 92.7 & 32.3 & 171 \\
\bottomrule
\end{tabular}}
\end{table}

%% file: sections/05_discussion.tex
\section{Discussion}
\label{sec:discussion}

The pipeline reduces data on both links. For a $100$\,km$^2$ urban area
($\approx$ Paris proper) at $0.8$\,m/px, full-resolution pre-disaster references
would weigh $\approx450$\,MB, against under $10$\,MB for $\times64$-compressed
latents; on the downlink, only compact detection products are transmitted instead
of full scenes. At the measured throughput ($\sim27$--$85$\,Mpx/s), inference
over such an area takes a few seconds: the downlink, not compute, is the
bottleneck, which is exactly what on-board product compression addresses.
On robustness, our results clearly separate the two targets. The alignment module is
the best answer to de-registration (flat mAP across the operational shift
range and the best nominal accuracy) but relies on grid sampling, a GPU
operator absent from the Versal DPU. Classical cross-attention is next best on
robustness yet equally non-portable (spatial softmax). The DPU-compatible
alternatives (a portable spatial-difference gate, and compression) keep the
model deployable but do \emph{not} recover long-range robustness. Thus, on a fixed-function DPU,
on-board \emph{correction} of large de-registration remains open, whereas an
embedded GPU (Jetson) supports the full robust pipeline today.

\paragraph{Limitations.}
This work is a step \emph{toward} raw on-board data, not a validation on it. We
operate on L1C-like inputs (ortho-rectified, calibrated, only coarsely
co-registered RGB pairs, as in xBD~\cite{gupta2019xbddatasetassessingbuilding})
and move closer to raw acquisitions by removing the need for accurate
co-registration. Part of the real variability is already exercised by xBD:
off-nadir differences surface as a fixed per-image pixel
drift~\cite{gupta2019xbddatasetassessingbuilding}, exactly the global shift our
alignment targets, and illumination and seasonal changes span nineteen events over
several years, though each factor is not evaluated separately. Remaining gaps: on-board sensor
calibration, scale and rotation, cross-sensor pairs and raw
acquisitions.

\paragraph{Perspectives.}
The shift estimation module is, in essence, a latent-space registration step: it
estimates the transform between the latent of the acquired image and that of a
reference encoded on the ground. Extended to scale and rotation, it would register
raw acquisitions on board against the compressed reference. On the DPU, which lacks grid
sampling, alternative alignment operators such as
TriCCOT~\cite{triccot} are being explored. Finally, as the pre-disaster branch
runs on the ground, a larger encoder could enrich the reference at no on-board
cost, and multi-temporal or multimodal (e.g.\ SAR) inputs would improve
robustness to illumination and weather.

%% file: sections/06_conclusion.tex
\section{Conclusion}
\label{sec:conclusion}

We presented an on-board data-reduction architecture for bi-temporal building
damage assessment that compresses pre-disaster information on the ground and
extracts compact damage products on board. Building on our previous
work~\cite{Goudemant_2026_CVPR}, it addresses three challenges of this
distributed ground/on-board use case. First, it cuts the uplink data volume by
about two orders of magnitude (up to $\times64$) with no notable loss in
detection, making the pre-disaster reference practical to transmit. Second, it
moves closer to raw on-board inputs: a latent-space shift estimation and
correction module regresses the pre/post offset and realigns the post-disaster
features before fusion, removing the need for accurate co-registration and
outperforming shift augmentation and cross-attention under strong de-registration.
Third, it is
deployed on embedded hardware (to our knowledge, the first on-board, dual
ground/on-board detector of its kind) on a Versal VCK190 and a Jetson
AGX~Orin. The core detector and its compression port cleanly, while the operators
giving long-range robustness are non-portable on the fixed-function DPU, whereas
the Jetson GPU runs the full pipeline. Beyond damage assessment, extending the latent shift estimation to scale and
rotation opens the way to registering raw acquisitions on board against a
compressed ground reference.